\documentclass[preprint]{elsarticle}

\usepackage{graphicx}
\usepackage{amssymb}

\usepackage{lineno}

\usepackage{amsmath}
\usepackage{booktabs}
\usepackage{multirow}

\journal{Journal Name}

\begin{document}

\begin{frontmatter}


\title{Comparison of RNN Encoder-Decoder Models for Anomaly Detection}



\author{YeongHyeon Park$^a$}
\ead{yeonghyeon@hufs.ac.kr}
\author{Il Dong Yun\corref{cor1}$^b$}
\ead{yun@hufs.ac.kr}
\cortext[cor1]{Corresponding author}
\address{$^{a,b}$Department of Computer and Electronic Systems Engineering,\\Hankuk University of Foreign Studies, Yongin, South Korea}

\begin{abstract}
In this paper, we compare different types of Recurrent Neural Network (RNN) Encoder-Decoders in anomaly detection viewpoint. We focused on finding the model that can learn the same data more effectively. We compared multiple models under the same conditions, such as the number of parameters, optimizer, and learning rate. However, the difference is whether to predict the future sequence or restore the current sequence. We constructed the dataset with simple vectors and used them for the experiment. Finally, we experimentally confirmed that the model performs better when the model restores the current sequence, rather than predict the future sequence.
\end{abstract}


\end{frontmatter}


\section{Introduction}
\label{sec:Intoroduction}
Currently, the Recurrent Neural Network (RNN) used to generating sequence based on the input sequence. The sequence generation model based on RNN generally called sequence to sequence RNN or RNN Encoder-Decoder (RED) \citep{sutskever2014sequence, cho2014learning}. In this paper, our purpose is finding the model that can learn the same data more effectively. We assumed the RED will be used for anomaly detection such as previous researches \citep{chauhan2015anomaly, malhotra2015long, nanduri2016anomaly}. In their research, they were presented the model of predicting the future sequence from the current sequence. We wondered whether it is the best way to using future prediction model for anomaly detection. Because, we think restoring the current sequence is easier than predicting the future.

We compared three RED models. The first model is the future prediction model. The second model is modified model of the future prediction model to restoring current sequence. The last model is an intermediate model of the previous two models. The detail of each model is described in Section \ref{sec:RED}. In Section \ref{sec:Experiment}, we present our dataset and experimental result and conclude in the final section.

\section{RNN Encoder-Decoder}
\label{sec:RED}
The RNN Encoder-Decoder (RED) have similarities with Auto-Encoder (AE) that generate the output from the inputs. The RED used to predict the next sequences in general. For example, Sucheta Chauhan et al. used RED for anomaly detection in Electrocardiography (ECG) \citep{chauhan2015anomaly}. Their RED was trained to predict the next (future) $L$ sequences from the input sequences. The ECG consists by $P$, $Q$, $R$, $S$ and $T$ waveforms with repeating. Each waveform has the one-to-one correspondence in ECG. That means only the wave $Q$ wave can appear after the $P$. So, there is no problem for training or predicting with previous RED. However, if the next wave of wave $P$ is not only $Q$ but also $R$, $S$ and $T$ appears directly (one-to-many correspondence), it makes confusing to learning. The RED derives the output by causality shown in Equation \ref{eq:act_input} to \ref{eq:hiddn_state}. When the model receives input sequence that starting with $P$, RED becomes confused to select output among various options. Because the first vector of the output sequence is determined only by the first vector of the input sequence without the prior information. We experimented with the above attributes of RED.

\begin{equation}
    i_{t} = \sigma(W_{i}x_{t} + U_{i}h_{t-1} + b_{i})
    \label{eq:act_input}
\end{equation}
\begin{equation}
    o_{t} = \sigma(W_{o}x_{t} + U_{o}h_{t-1} + b_{o})
    \label{eq:act_output}
\end{equation}
\begin{equation}
	f_{t} = \sigma(W_{f}x_{t} + U_{f}h_{t-1} + b_{f})
    \label{eq:act_forget}
\end{equation}
\begin{equation}
    c_{t} = f_{t}\circ c_{t-1} + i_{t}\circ tanh(W_{c}x_{t}+U_{c}h_{t-1}+b_{c})
    \label{eq:cell_state}
\end{equation}
\begin{equation}
    h_{t} = o_{t}\circ tanh(c_{t})
	\label{eq:hiddn_state}
\end{equation}

\subsection{Models for comparison}
\label{subsec:Modelsforcomparison}
In the RED, each RNN cell can be selectively used among vanilla RNN \citep{mikolov2010recurrent}, Long-short Term Memory (LSTM) \citep{hochreiter1997long} and Gated Recurrent Unit (GRU) \citep{chung2015gated}. The vanilla RNN has vanishing gradient problem when the length of the sequence becomes long. However, LSTM and GRU solved vanishing gradient problem.

\begin{figure}[ht]
	\begin{center}
		\includegraphics[width=0.4\linewidth]{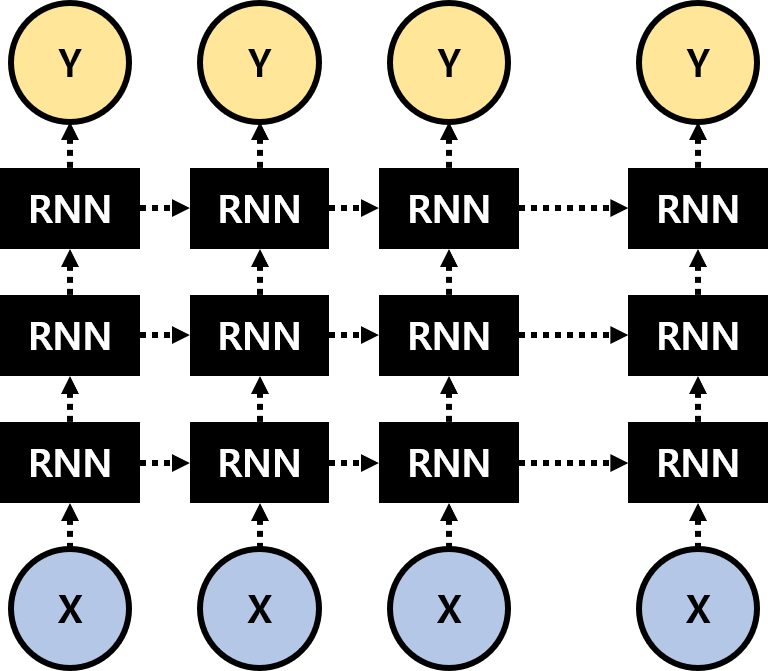}
	\end{center}
	\vspace*{-5mm}
	\caption{The RED architecture. It generates sequential output from sequential input.}
    \label{fig:red_architecture}
\end{figure}

We construct REDs using LSTM shown as Figure \ref{fig:red_architecture}. Because, there is no superiority between LSTM and GRU that Chung et al. have already confirmed \citep{chung2014empirical}. The relationship of input and output sequence can be adjusted for purpose. The relationship between the input and output was current $L$ sequence and future $L$ sequence in previous researches. In that case, RED learned to generate (predict) the future sequence from the current sequence. 

\begin{equation}
    p(\mathbf{Y}|\mathbf{X}) \propto{p(\mathbf{X}|\mathbf{Y})p(\mathbf{Y})}
    \label{eq:cond_prob}
\end{equation}

In the Equation \ref{eq:cond_prob}, $\mathbf{X}$ and $\mathbf{Y}$ are input and output sequence respectively. The $p(\mathbf{Y}|\mathbf{X})$ is the uncertainty, $p(\mathbf{X}|\mathbf{Y})$ is likelihood and $p(\mathbf{Y})$ is prior knowledge. The RED will be learned $p(\mathbf{X}|\mathbf{Y})$ by the maximum likelihood estimation in the training process. If $\mathbf{X}$ and $\mathbf{Y}$ are almost the same, it will be easier to learn likelihood, but vice versa. Because, it is easier to get the $A'$ than to get $B$ from the information $A$.

\begin{figure}[ht]
	\begin{center}
		\begin{tabular}{c}
            \includegraphics[width=0.7\linewidth]{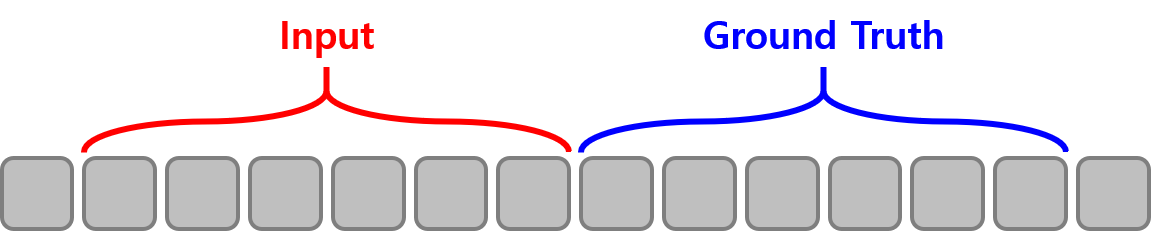} \\
            \includegraphics[width=0.7\linewidth]{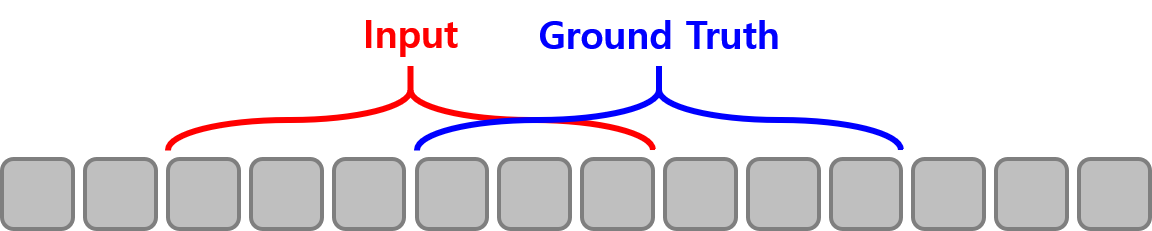} \\
            \includegraphics[width=0.7\linewidth]{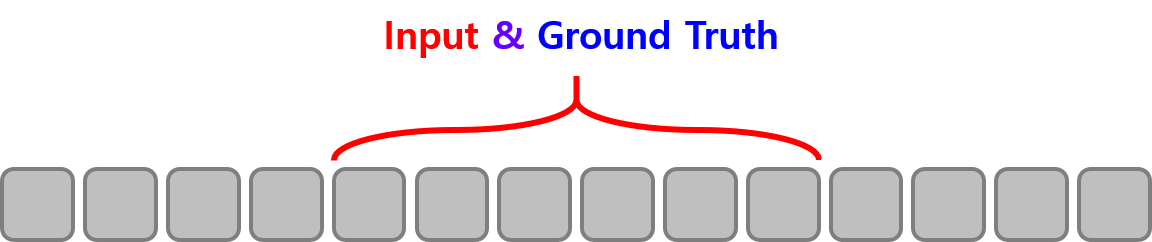}
		\end{tabular}
	\end{center}
	\vspace*{-5mm}
	\caption{The RED models for comparison. We call these Model-A, Model-B and Model-C from the top.}
	\label{fig:models}
\end{figure}

We construct three models for experiment. Model-A is future prediction model that same as the model in the previous researches \citep{chauhan2015anomaly, malhotra2015long, nanduri2016anomaly}. Model-C is restoration model that generates output sequence same as input sequence. Model-B is an intermediate model of Model-A and Model-C. We assume that the Model-C will be better to train when the dataset has one-to-many correspondence as mentioned earlier.

\pagebreak
\section{Experiment}
\label{sec:Experiment}
In this section, we experimentally compare the three RED models. We constructed the simple sequential dataset for the experiment that described in Section \ref{subsec:Dataset}.

\subsection{Dataset}
\label{subsec:Dataset}
We constructed the dataset with 15 simple vectors. We called each vector $A$ (Alpha) to $O$ (Oscar) in North Atlantic Treaty Organization (NATO) phonetic alphabet and converted it to the one-hot encoded vector shown as Equation \ref{eq:onehot_alphabet}.

\begin{equation}
	\begin{bmatrix} 
    A \\
    B \\
    C \\
    \vdots \\
    O 
    \end{bmatrix}
    =
	\begin{bmatrix} 
    1 & 0 & 0 & 0 & \dots & 0 \\
    0 & 1 & 0 & 0 & \dots & 0 \\
    0 & 0 & 1 & 0 & \dots & 0 \\
    \vdots &  &  &  & \ddots & \vdots \\
    0 & 0 & 0 & 0 & \dots & 1 \\
    \end{bmatrix}
    \label{eq:onehot_alphabet}
\end{equation}

We constructed three datasets with 15 vectors shown in Equation \ref{eq:onehot_alphabet}. One of the datasets consists of 5 vectors between $A$ to $E$ (Echo) and other datasets are consisting of whole 15 vectors. The pattern of each dataset is shown in Table \ref{table:datasets}. The patterns of each dataset are shown in Table \ref{table:datasets} repeated for a length of 3000. The notable point of three datasets in Table \ref{table:datasets} is Alpha symbol of Set-C. The symbol Alpha appears in front of symbol $B$ (Bravo), $D$ (Delta), $F$ (Foxtrot), $H$ (Hotel) and $J$ (Juliett). That makes confusing when training the pattern to RED because RED cannot generate the output correctly when the sequence started with Alpha.

\begin{table}[h]
    \caption{Three datasets for training process. Each pattern is repeating itself.}
    \centering
    \small
        \begin{tabular}{llr}
            \toprule
            \textbf{Dataset} & \textbf{Pattern} & \textbf{Length} \\
            \midrule
            Set-A   & ABCDE & 5 \\
            Set-B   & ABCDEFGHIJKLMNO & 15 \\
            Set-C   & ABCADEAFGAHIAJK & 15 \\
            \bottomrule
        \end{tabular}
    \label{table:datasets}
\end{table}

Each dataset is subdivided into four types. One of subset consists of the same pattern as Table \ref{table:datasets} and another one is added random noise to the original pattern with. Also, the dataset contains the new pattern that has not seen in Table \ref{table:datasets} and added noise to them. We call each subset as 'Clear', 'Noise', 'Abnormal' and 'Abnoise' (abnormal with noise). Subset 'Clear' and 'Noise' is normal class and others are abnormal class. We use only one normal class subset of the all subsets for training. The test process uses all subsets.

\pagebreak
\subsection{Comparison}
\label{subsec:Comparison}
We compared the three models with three datasets. We trained each model with subset 'Clear' and 'Noise' respectively. We defined the synchronous case when the length of the pattern and the sequence length are equal or integer multiple. And other cases to asynchronous. We experimented with three synchronous cases and three asynchronous cases. Loss graphs of training process is shown in \ref{adx:a}. We evaluated three models with training set to confirm they learned well about the training set. Table \ref{table:train_loss_clear} and \ref{table:train_loss_noise} shows the result of evaluation. The lower loss means the model generates output more correctly. Model-C has lower loss than other models for all the training set.

\begin{table}[h]
	\caption{Loss matrix with training set. Each model trained with subset 'clear'.}
    \centering
    \small
    \begin{tabular}{crrr}
		\toprule
        \textbf{Dataset (Subset)} & \textbf{Model-A} & \textbf{Model-B} & \textbf{Model-C} \\
        \midrule
        Set-A (Clear) & 3.291 & 3.440 & 2.484 \\
        Set-B (Clear) & 2.680 & 2.974 & 2.377 \\
        Set-C (Clear) & 2.851 & 3.019 & 2.672 \\
        \bottomrule
    \end{tabular}
    \label{table:train_loss_clear}
\end{table}

\begin{table}[h]
	\caption{Loss matrix with training set. Each model trained with subset 'noise'.}
    \centering
    \small
    \begin{tabular}{crrr}
		\toprule
        \textbf{Dataset (Subset)} & \textbf{Model-A} & \textbf{Model-B} & \textbf{Model-C} \\
        \midrule
        Set-A (Noise) & 5.253 & 5.380 & 4.771 \\
        Set-B (Noise) & 4.615 & 4.888 & 4.378 \\
        Set-C (Noise) & 4.781 & 4.844 & 4.499 \\
        \bottomrule
    \end{tabular}
    \label{table:train_loss_noise}
\end{table}

We also evaluated three models with the test set shown as Table \ref{table:test_loss_clear} and \ref{table:test_loss_noise}. Tables are presented the average loss for each class and model. The loss was measured by a divided normal set and abnormal set in each the test set. The lower loss for the normal set is better same as evaluation with the training set and for the abnormal set is vice versa.

\begin{table}[h]
	\caption{Loss matrix with test set. Each model trained with subset 'Clear'.}
    \centering
    \small
    \begin{tabular}{clrrr}
		\toprule
        \textbf{Dataset} & \textbf{Class} & \textbf{Model-A} & \textbf{Model-B} & \textbf{Model-C} \\
        \midrule
        \multirow{2}{*}{Set-A} & Normal   & 4.046 & 4.184 & 3.339 \\
                               & Abnormal & 5.180 & 5.238 & 4.484 \\
        \midrule
        \multirow{2}{*}{Set-B} & Normal   & 3.482 & 3.698 & 3.170 \\
                               & Abnormal & 4.351 & 4.554 & 3.955 \\
        \midrule
        \multirow{2}{*}{Set-C} & Normal   & 3.534 & 3.704 & 3.392 \\
                               & Abnormal & 4.929 & 4.905 & 4.116 \\
        \bottomrule
    \end{tabular}
    \label{table:test_loss_clear}
\end{table}

\begin{table}[h]
	\caption{Loss matrix with test set. Each model trained with subset 'Noise'.}
    \centering
    \small
    \begin{tabular}{clrrr}
		\toprule
        \textbf{Dataset} & \textbf{Class} & \textbf{Model-A} & \textbf{Model-B} & \textbf{Model-C} \\
        \midrule
        \multirow{2}{*}{Set-A} & Normal   & 4.012 & 4.198 & 3.452 \\
                               & Abnormal & 5.169 & 5.353 & 4.571 \\
        \midrule
        \multirow{2}{*}{Set-B} & Normal   & 3.544 & 3.825 & 3.285 \\
                               & Abnormal & 4.365 & 4.603 & 3.285 \\
        \midrule
        \multirow{2}{*}{Set-C} & Normal   & 3.691 & 3.752 & 3.392 \\
                               & Abnormal & 4.961 & 4.874 & 4.103 \\
        \bottomrule
    \end{tabular}
    \label{table:test_loss_noise}
\end{table}

\pagebreak
Since confirming the superiority only with the loss value is difficult, we also checked the generated sequence. We present the part of the test case in Table \ref{table:seta_seq3_clear} and \ref{table:setc_seq8_clear}. We provide the whole test result in our Github repository \footnote{https://github.com/YeongHyeon/Compare\_REDs}.

\begin{table}[h]
	\caption{The part of test result with Set-A. The sequence length is 3 (asynchronous case) and each model trained with subset 'Clear'.}
    \centering
    \small
    \begin{tabular}{cllllr}
    	\toprule
        \textbf{Model} & \textbf{Subset} & \textbf{Input} & \textbf{Output} & \textbf{Ground-Truth} & \textbf{Loss} \\
        \midrule
        \multirow{4}{*}{Model-A} & Clear    & ABC & DEA & DEA & 0.244 \\
                                 & Noise    & DEA & BCD & BCD & 1.480 \\
                                 & Abnormal & BDE & EAB & ACC & 2.106 \\
                                 & Abnoise  & BBC & EAA & DEB & 1.926 \\
        \midrule
        \multirow{4}{*}{Model-B} & Clear    & ABC & BCD & BCD & 0.244 \\
                                 & Noise    & DEA & EAB & EAB & 1.104 \\
                                 & Abnormal & BDE & CDA & DEA & 1.794 \\
                                 & Abnoise  & BBC & CDE & BCD & 2.283 \\
        \midrule
        \multirow{4}{*}{Model-C} & Clear    & ABC & ABC & ABC & 0.244 \\
                                 & Noise    & DEA & DEA & DEA & 1.261 \\
                                 & Abnormal & BDE & BCD & BDE & 1.707 \\
                                 & Abnoise  & BBC & BCC & BBC & 1.747 \\
        \bottomrule
    \end{tabular}
    \label{table:seta_seq3_clear}
\end{table}

\begin{table}[h]
	\caption{The part of test result with Set-C. The sequence length is 8 (asynchronous case) and each model trained with subset 'Clear'.}
    \centering
    \small
    \begin{tabular}{cllllr}
    	\toprule
        \textbf{Model} & \textbf{Class} & \textbf{Input} & \textbf{Output} & \textbf{Ground-Truth} & \textbf{Loss} \\
        \midrule
        \multirow{4}{*}{Model-A} & Clear    & ABCADEAF & EAHIAJKA & GAHIAJKA & 0.913 \\
                                 & Noise    & AFGAHIAJ & DABCADEA & KABCADEA & 2.169 \\
                                 & Abnormal & JKBBBADE & AFGAHIAJ & AFGAHIAJ & 0.243 \\
                                 & Abnoise  & AKKKBCAD & DAFGAHIA & EAFGAHIK & 2.468 \\
        \midrule
        \multirow{4}{*}{Model-B} & Clear    & ABCADEAF & BEAFGAHI & DEAFGHAI & 0.917 \\
                                 & Noise    & AFGAHIAJ & JIAJKABC & HIAJKAVC & 2.100 \\
                                 & Abnormal & JKBBBADE & CADEAAGA & BADEAFGA & 1.944 \\
                                 & Abnoise  & AKKKBCAD & JAAAEAJA & BCADEAFG & 3.119 \\
        \midrule
        \multirow{4}{*}{Model-C} & Clear    & ABCADEAF & ABCADEAF & ABCADEAF & 0.241 \\
                                 & Noise    & AFGAHIAJ & AFGAHIAJ & AFGAHIAJ & 1.673 \\
                                 & Abnormal & JKBBBADE & JKABBADE & JKBBBADE & 1.268 \\
                                 & Abnoise  & AKKKBCAD & AKAKBCAD & AKKKBCAD & 2.818 \\
        \bottomrule
    \end{tabular}
    \label{table:setc_seq8_clear}
\end{table}

Each model generates the output sequences correctly when it received normal sequences, but it cannot generate the output correctly when it received abnormal sequences as shown in Table \ref{table:seta_seq3_clear}. We use this property for anomaly detection. However, we confirmed in Table \ref{table:setc_seq8_clear}, Model-A did not correctly predict the next sequence when using Set-C whether the input was normal or abnormal. Model-A may be performing a high recall for anomaly detection, but it can be always decided the state to abnormal. Model-C correctly restore the input sequence as output when it receipted normal sequences.

\begin{table}[h]
	\caption{The part of test result with Set-C. The sequence length is 8 (asynchronous case) and each model trained with subset 'Noise'.}
    \centering
    \small
    \begin{tabular}{cllllr}
    	\toprule
        \textbf{Model} & \textbf{Class} & \textbf{Input} & \textbf{Output} & \textbf{Ground-Truth} & \textbf{Loss} \\
        \midrule
        \multirow{4}{*}{Model-A} & Clear    & ABCADEAF & CAHIAJKA & GAHIAJKA & 1.986 \\
                                 & Noise    & AFGAHIAJ & CABCADEA & KABCADEA & 0.858 \\
                                 & Abnormal & JKBBBADE & AFGAHIAJ & AFGAHIAJ & 1.794 \\
                                 & Abnoise  & AKKKBCAD & CFFGAHIA & EAFGAHIK & 2.006 \\
        \midrule
        \multirow{4}{*}{Model-B} & Clear    & ABCADEAF & JEAFGAHI & DEAFGHAI & 2.018 \\
                                 & Noise    & AFGAHIAJ & FIAJKABC & HIAJKAVC & 0.900 \\
                                 & Abnormal & JKBBBADE & CADEAFGA & BADEAFGA & 2.412 \\
                                 & Abnoise  & AKKKBCAD & FAAADAFG & BCADEAFG & 2.263 \\
        \midrule
        \multirow{4}{*}{Model-C} & Clear    & ABCADEAF & ABCADEAF & ABCADEAF & 1.996 \\
                                 & Noise    & AFGAHIAJ & AFGAHIAJ & AFGAHIAJ & 0.283 \\
                                 & Abnormal & JKBBBADE & JKABBIDE & JKBBBADE & 2.699 \\
                                 & Abnoise  & AKKKBCAD & AKKKACAD & AKKKBCAD & 1.213 \\
        \bottomrule
    \end{tabular}
    \label{table:setc_seq8_noise}
\end{table}

\pagebreak
Table \ref{table:setc_seq8_noise} shows part of the test results when each model trained with subset 'Noise'. Model-A and Model-B cannot predict the next sequences correctly when using the subset 'Clear' unlike Table \ref{table:setc_seq8_clear}, but Model-C is not. Model-A constructed for focused on predicting future sequences rather than current key features. However, since Model-C constructed to restore current sequences, it can concentrate on learning key features in a given pattern. So, Model-C can get the robustness whether noise is added to the sequence.

\section{Conclusion}
\label{sec:Conclusion}
We experimentally compared the three models of RED. We assume the RED are used for anomaly detection. The RED generates or predicts the next sequences in general. But we wondered whether the future predicting is the best way for anomaly detection. There is no difference in predicting future time $t$, determining anomalies, and restoring and immediately determining the current state. The only difference is the difficulty of the training process. 

The model selection and setting of the sequence length are not important when the sequential vector has one-to-one correspondence property. But that case is a very ideal, and it is almost rare. There is a probability that various pattern appears randomly after the specific pattern like Set-C in reality. It is possible to get the correct result by an irregular pattern like Set-C by accident, but it is usually difficult.

We presented in the Section \ref{subsec:Comparison}, Model-C learned the same data more easily and effectively than other models. We certain restoring the current sequence like Model-C is better than predicting the future sequence like Model-A and Model-C for anomaly detection. The smoother the training process means it can performing better, and we have confirmed it with experiment.

When using the Model-A for anomaly detection, it will face the problem of misjudging the normal state as abnormal. The Model-C does not make correct judgments absolutely, but at least it will perform much better than other models. Also, we expect the Model-C can be using other types of time series data generation problems and performing well.



\section{References}



\appendix

\section{}
\label{adx:a}
\ref{adx:a} contains loss graphs of training process. Each process was performed with three models and three datasets respectively. Left side of each figure presents the synchronous case and the right side is the asynchronous case. Firstly, we show the results of the training process with subset 'Clear'. Last three figures are results of the training process with subset 'Noise'.

\begin{figure}[h]
	\small
    \begin{center}
		\begin{tabular}{c c}
			\includegraphics[width=0.43\linewidth,keepaspectratio=true]{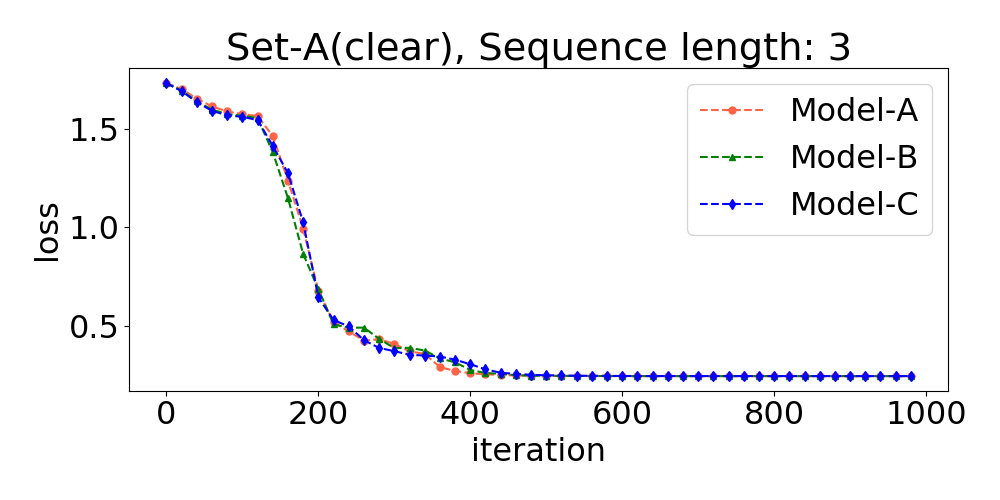} &
			\includegraphics[width=0.43\linewidth,keepaspectratio=true]{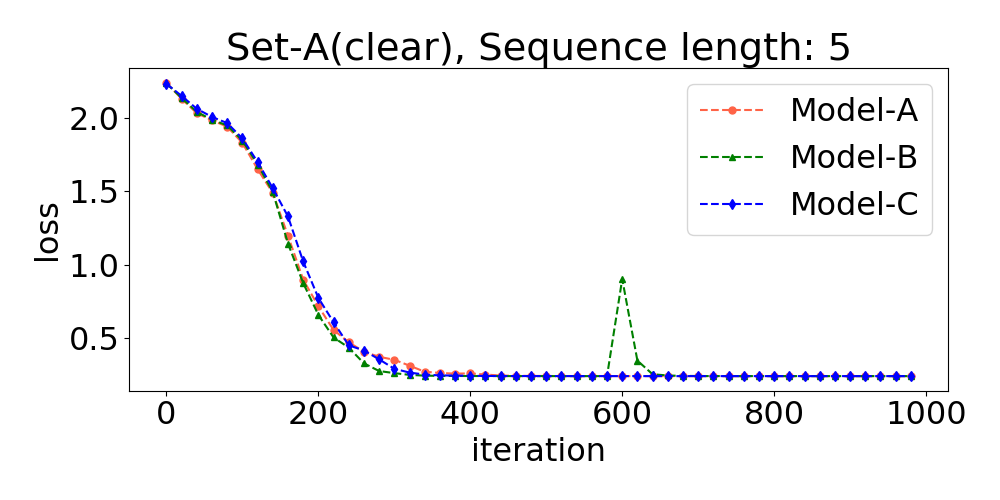} \\
            \includegraphics[width=0.43\linewidth,keepaspectratio=true]{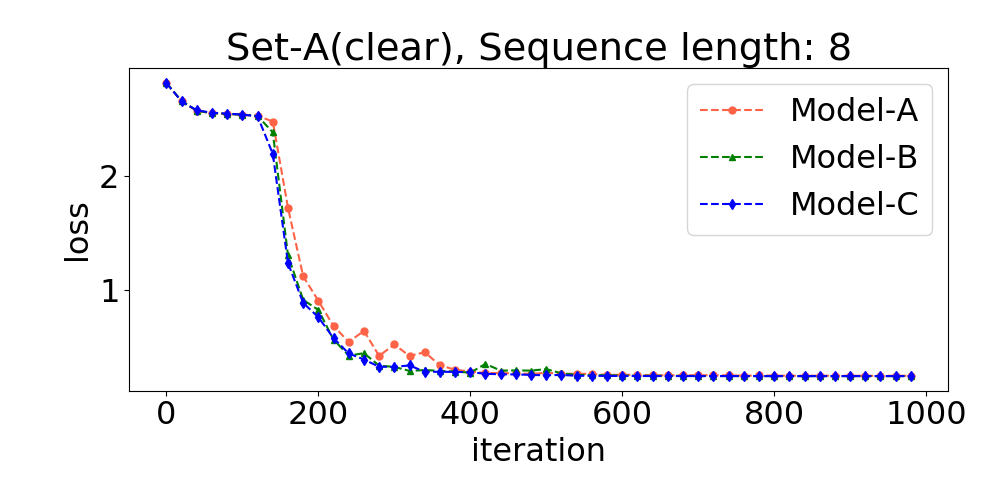} &
			\includegraphics[width=0.43\linewidth,keepaspectratio=true]{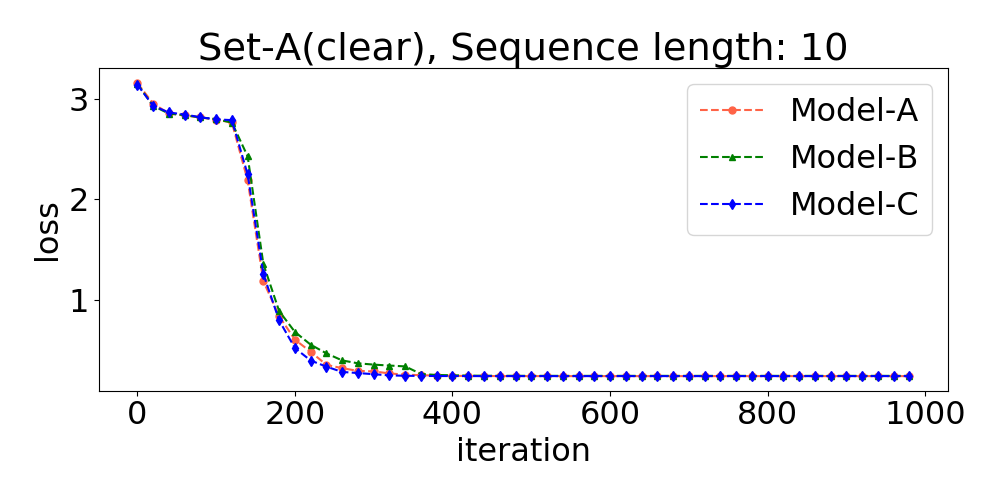} \\
            \includegraphics[width=0.43\linewidth,keepaspectratio=true]{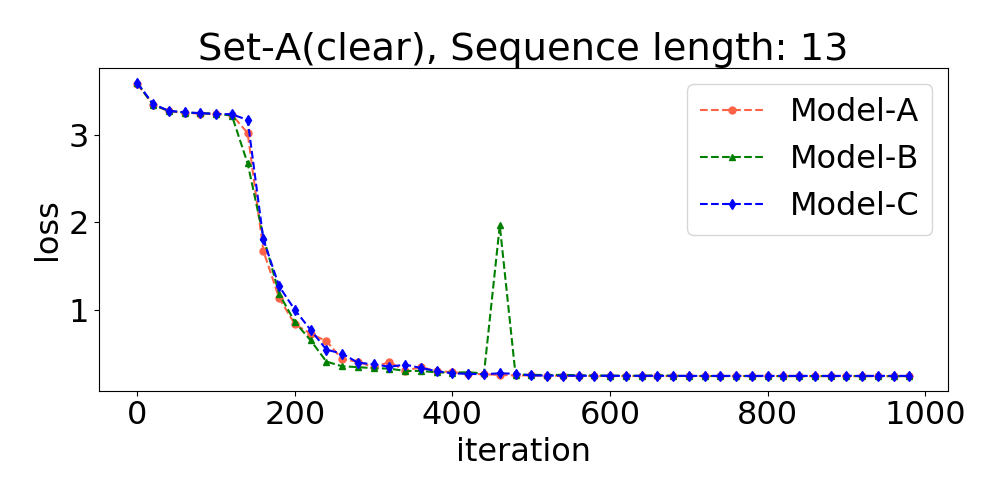} &
			\includegraphics[width=0.43\linewidth,keepaspectratio=true]{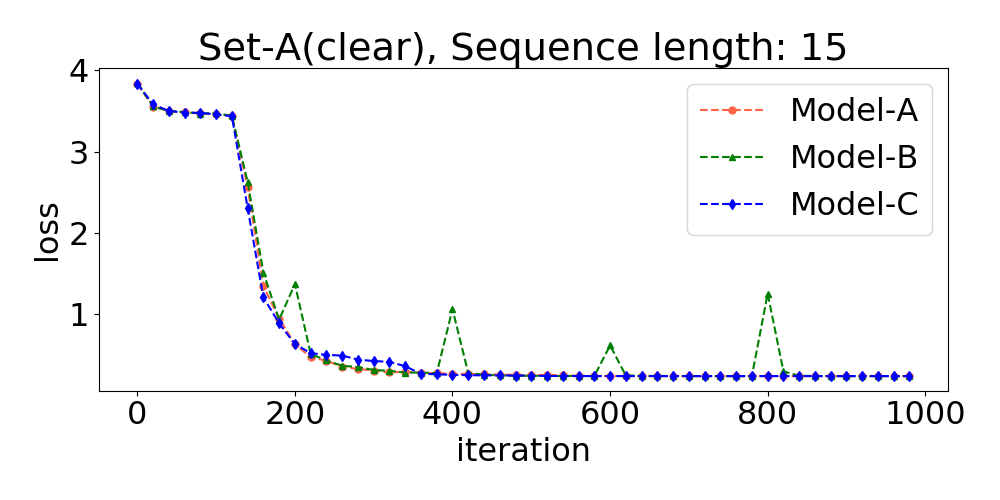} \\
		\end{tabular}
	\end{center}
    \caption{Loss graph of training process with subset 'Clear' of the Set-A. Loss value of each model converged almost similar moment at each sequence length.}
    \label{fig:mfcc_classes}
\end{figure}

\begin{figure}[h]
	\small
    \begin{center}
		\begin{tabular}{c c}
			\includegraphics[width=0.43\linewidth,keepaspectratio=true]{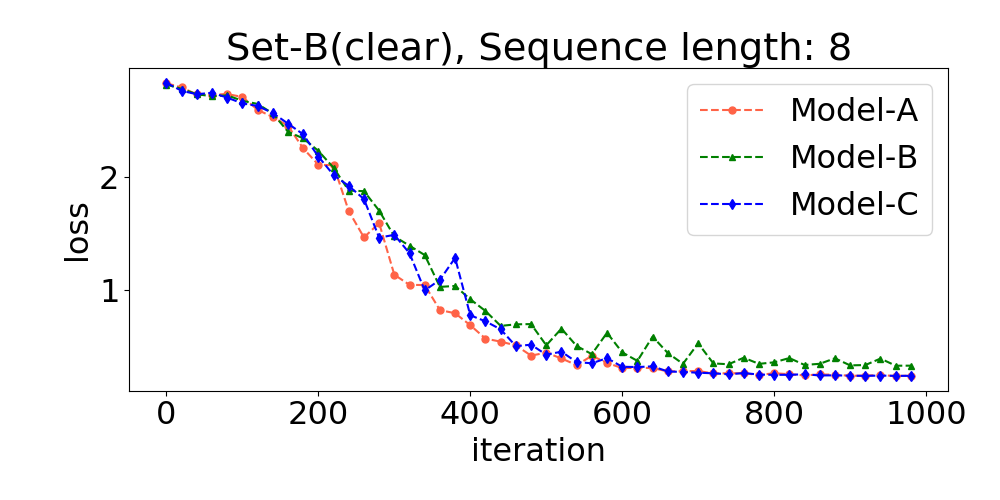} &
			\includegraphics[width=0.43\linewidth,keepaspectratio=true]{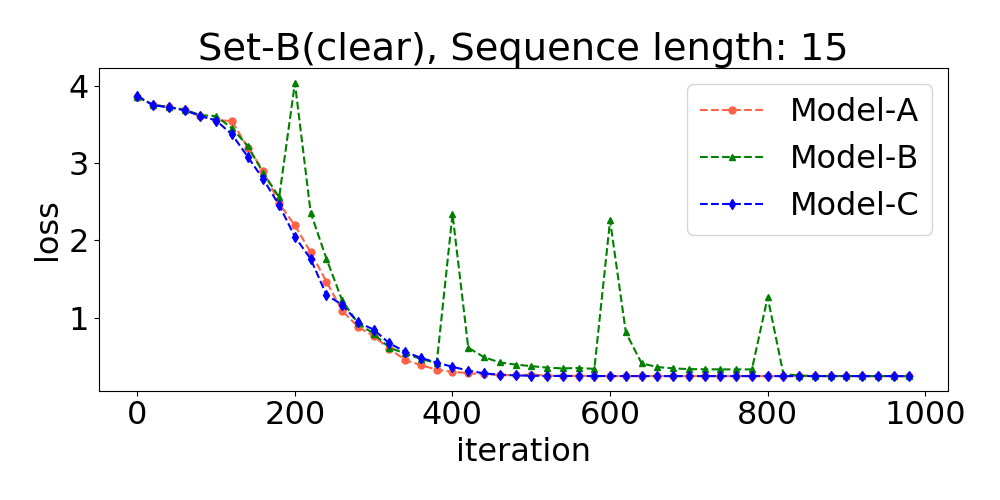} \\
            \includegraphics[width=0.43\linewidth,keepaspectratio=true]{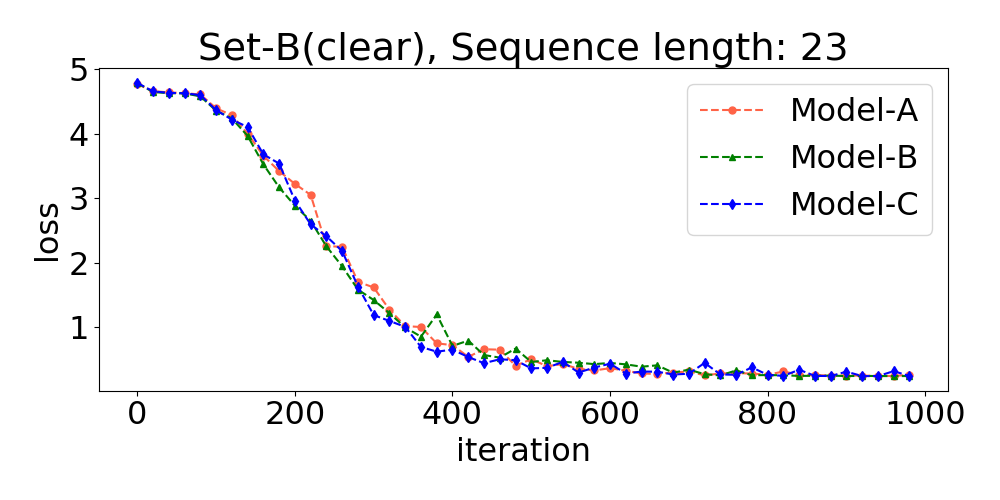} &
			\includegraphics[width=0.43\linewidth,keepaspectratio=true]{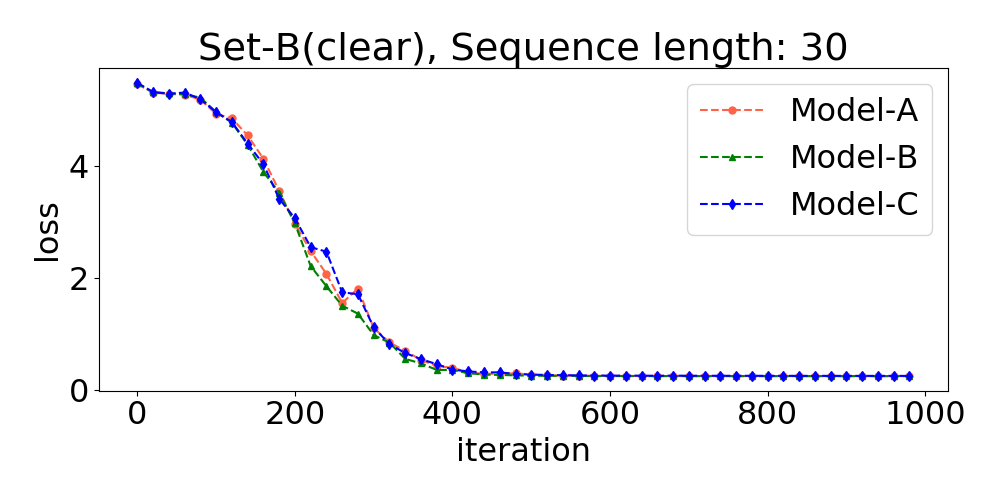} \\
            \includegraphics[width=0.43\linewidth,keepaspectratio=true]{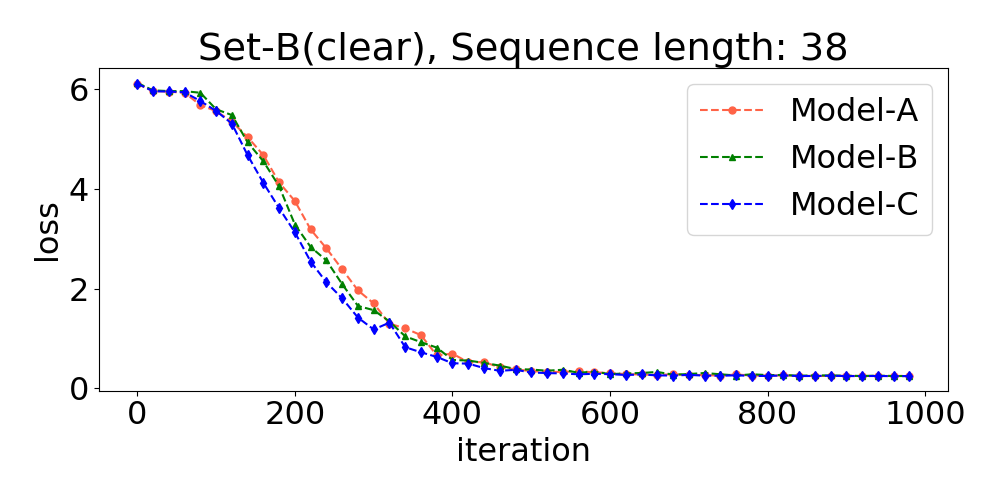} &
			\includegraphics[width=0.43\linewidth,keepaspectratio=true]{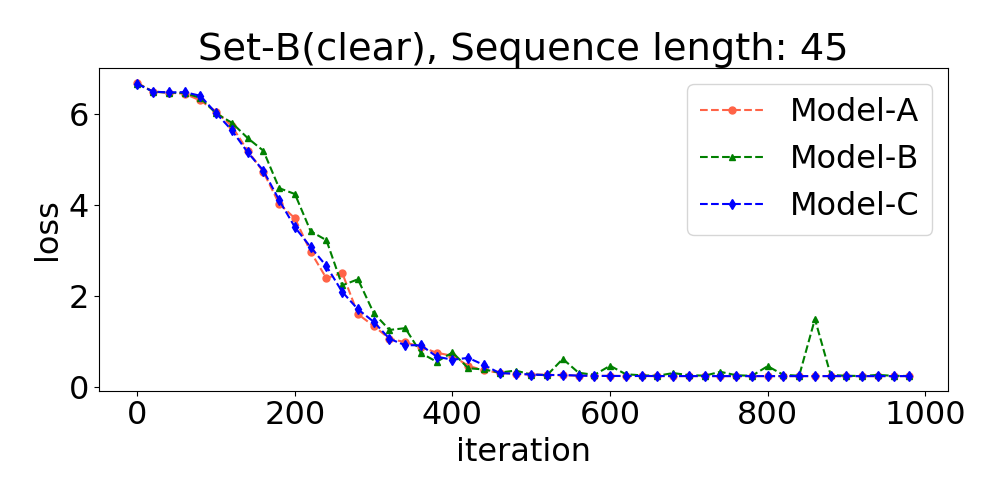} \\
		\end{tabular}
	\end{center}
    \caption{Loss graph of training process with subset 'Clear' of the Set-B.}
    \label{fig:mfcc_classes}
\end{figure}

\begin{figure}[h]
	\small
    \begin{center}
		\begin{tabular}{c c}
			\includegraphics[width=0.43\linewidth,keepaspectratio=true]{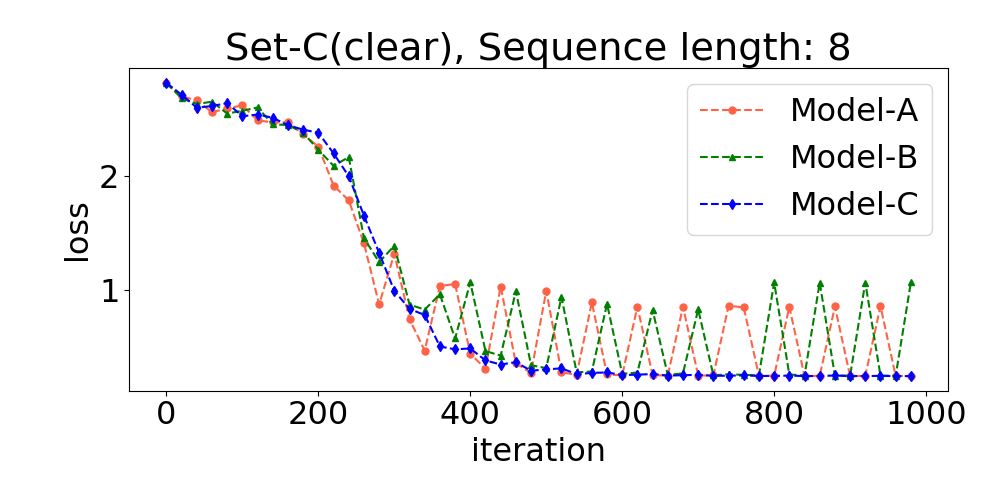} &
			\includegraphics[width=0.43\linewidth,keepaspectratio=true]{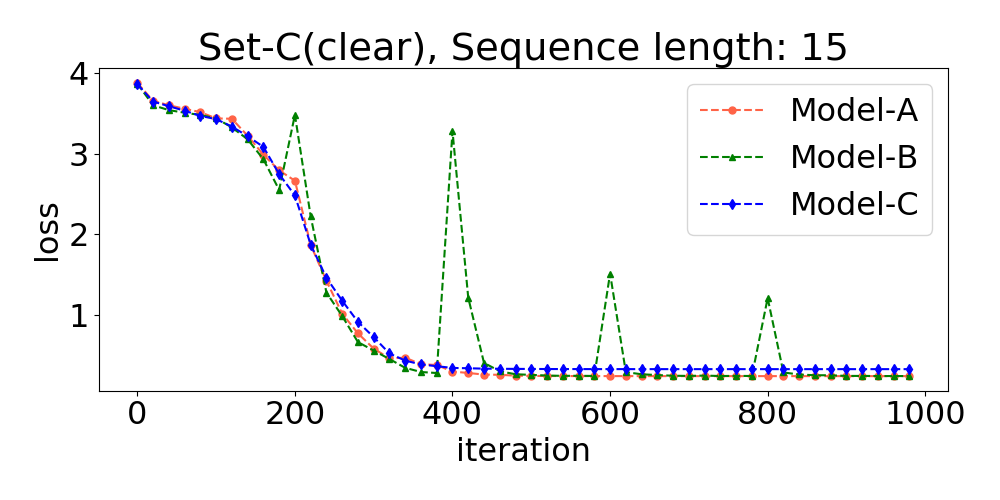} \\
            \includegraphics[width=0.43\linewidth,keepaspectratio=true]{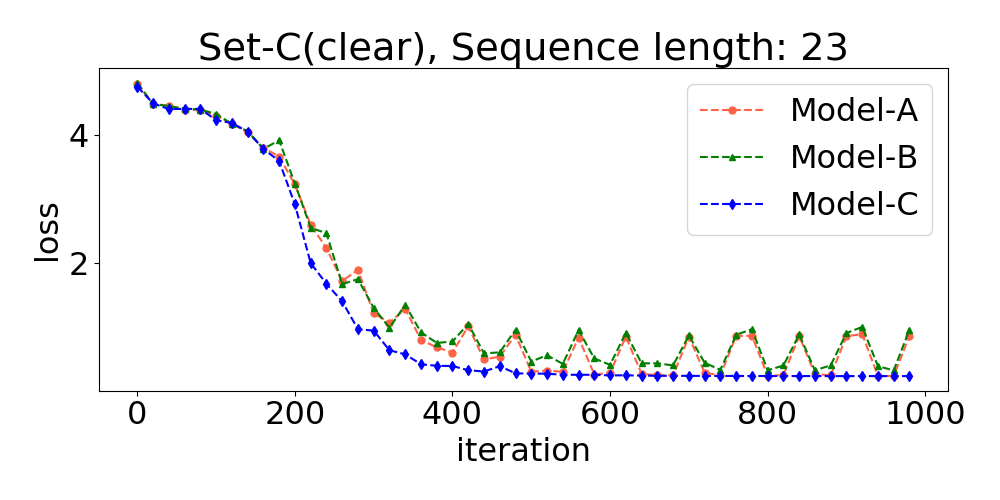} &
			\includegraphics[width=0.43\linewidth,keepaspectratio=true]{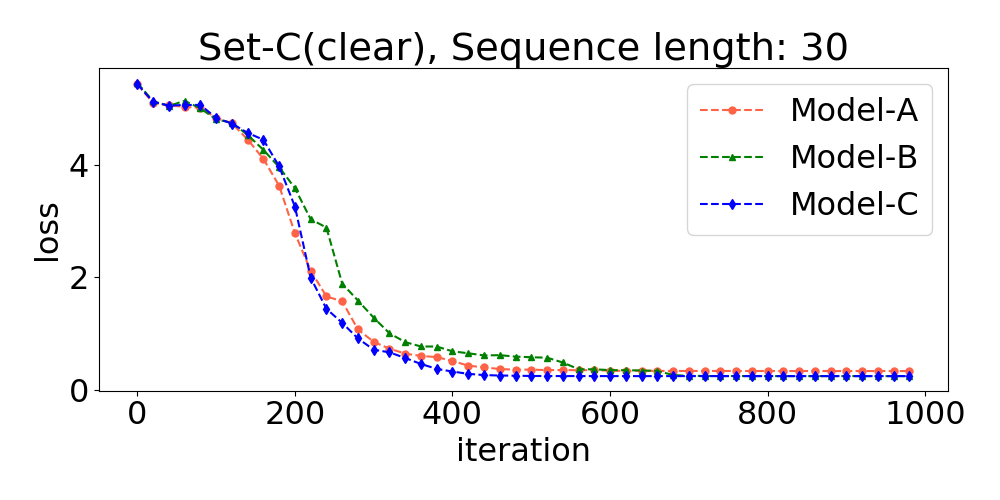} \\
            \includegraphics[width=0.43\linewidth,keepaspectratio=true]{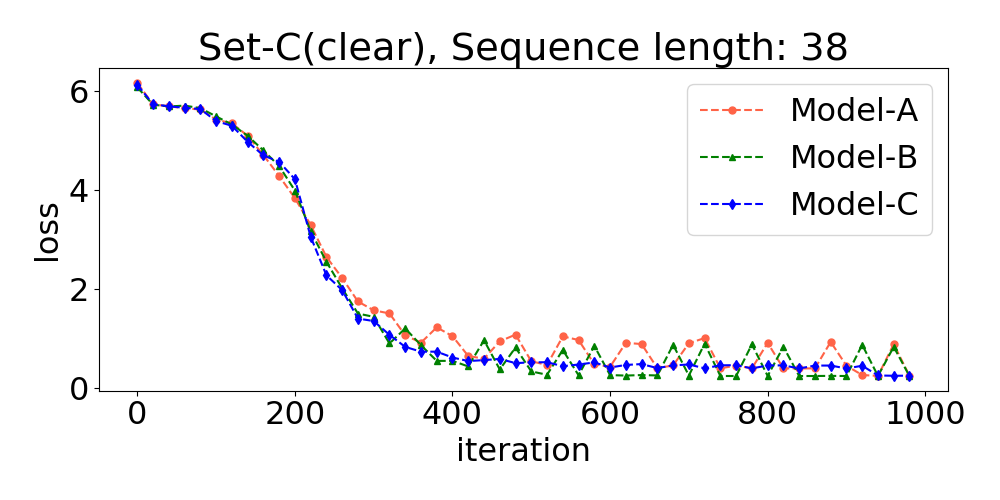} &
			\includegraphics[width=0.43\linewidth,keepaspectratio=true]{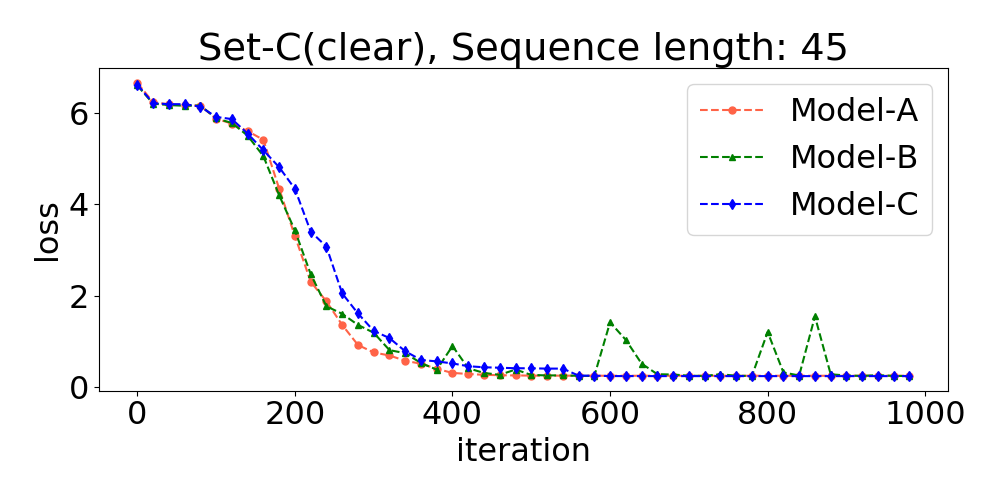} \\
		\end{tabular}
	\end{center}
    \caption{Loss graph of training process with subset 'Clear' of the Set-C. The loss value of Model-A and Model-B oscillated in asynchronous cases but Model-C is not.}
    \label{fig:mfcc_classes}
\end{figure}

\begin{figure}[h]
	\small
    \begin{center}
		\begin{tabular}{c c}
			\includegraphics[width=0.45\linewidth,keepaspectratio=true]{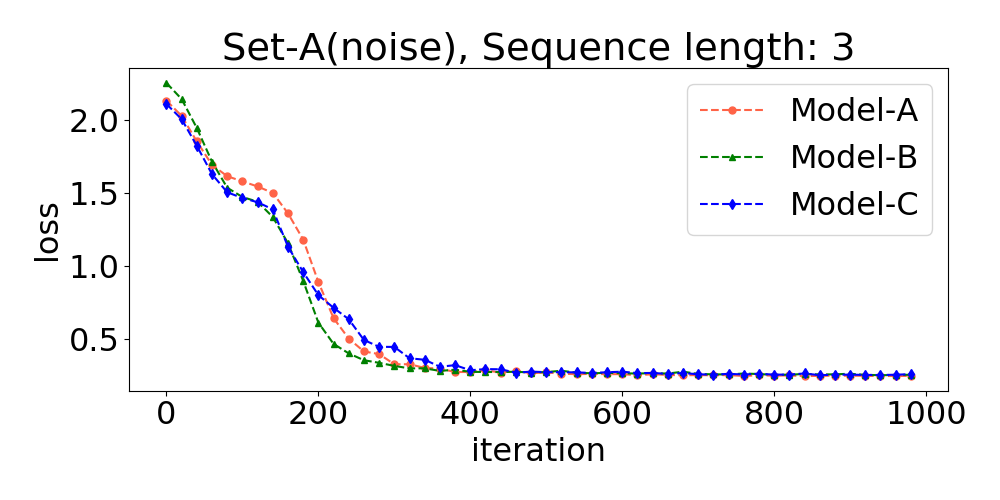} &
			\includegraphics[width=0.45\linewidth,keepaspectratio=true]{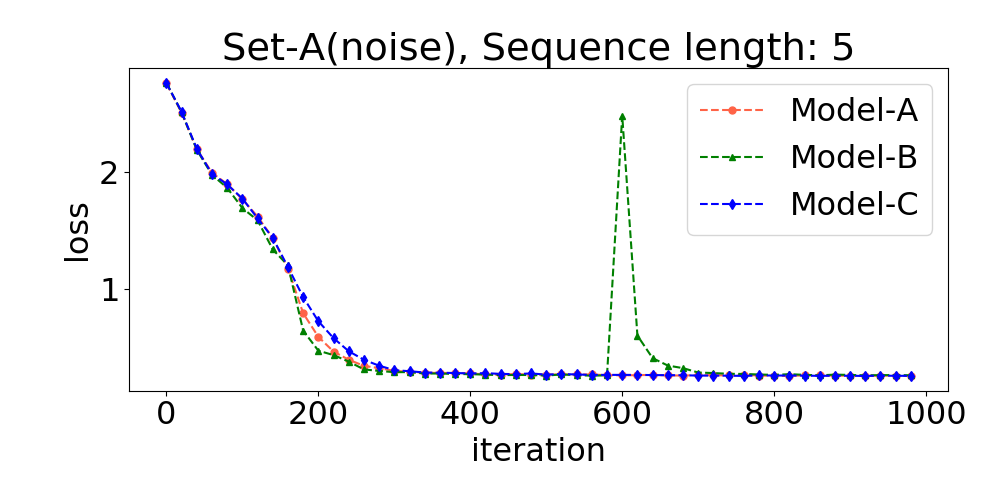} \\
            \includegraphics[width=0.45\linewidth,keepaspectratio=true]{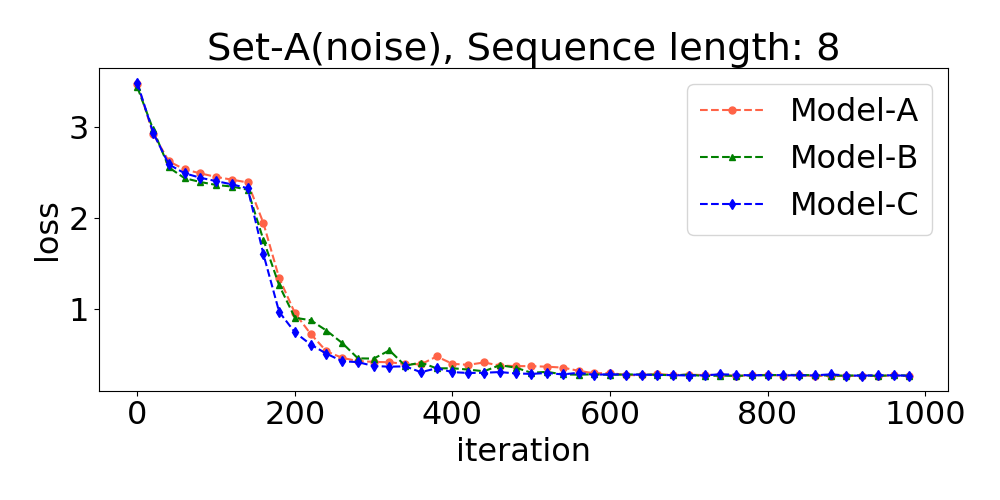} &
			\includegraphics[width=0.45\linewidth,keepaspectratio=true]{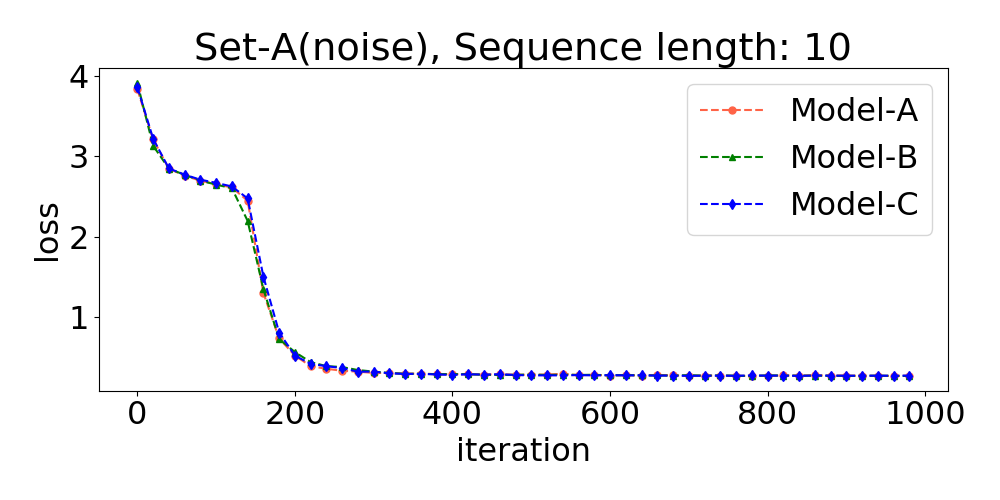} \\
            \includegraphics[width=0.45\linewidth,keepaspectratio=true]{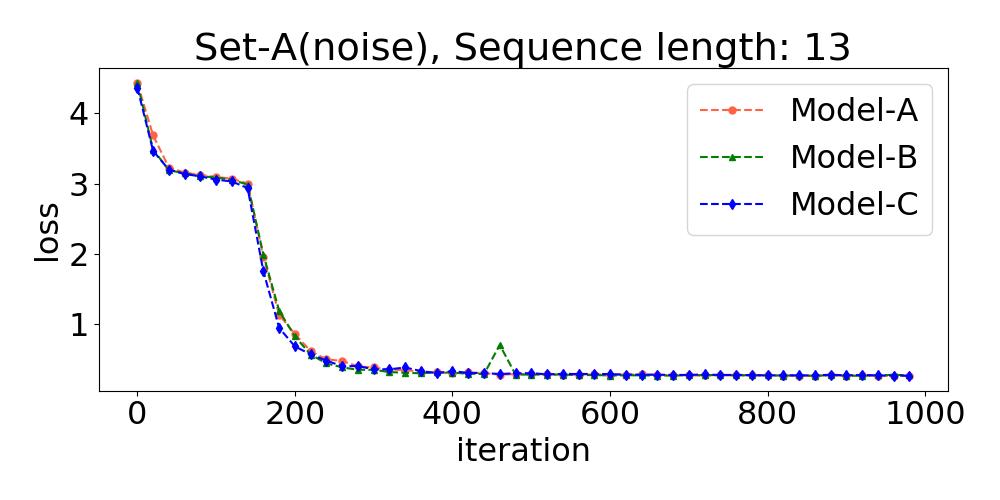} &
			\includegraphics[width=0.45\linewidth,keepaspectratio=true]{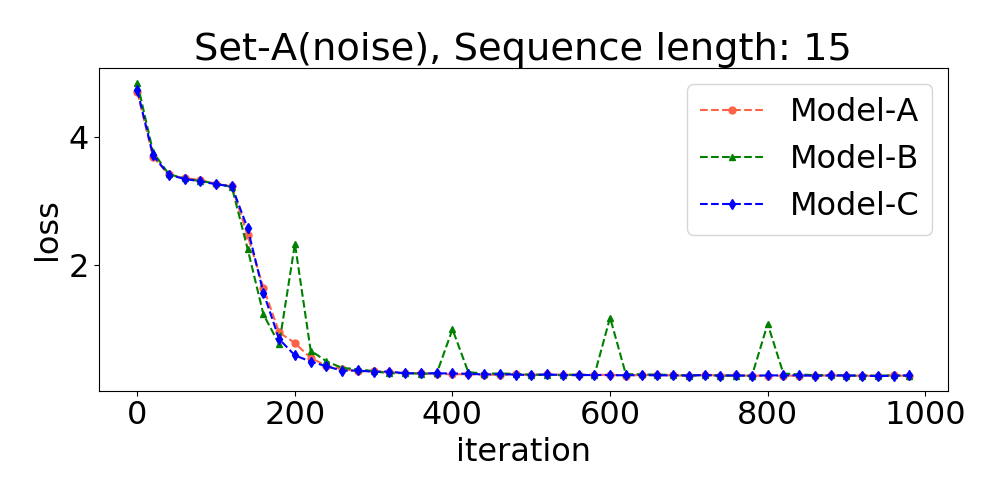} \\
		\end{tabular}
	\end{center}
    \caption{Loss graph of training process with subset 'Noise' of the Set-A.}
    \label{fig:mfcc_classes}
\end{figure}

\begin{figure}[h]
	\small
    \begin{center}
		\begin{tabular}{c c}
			\includegraphics[width=0.45\linewidth,keepaspectratio=true]{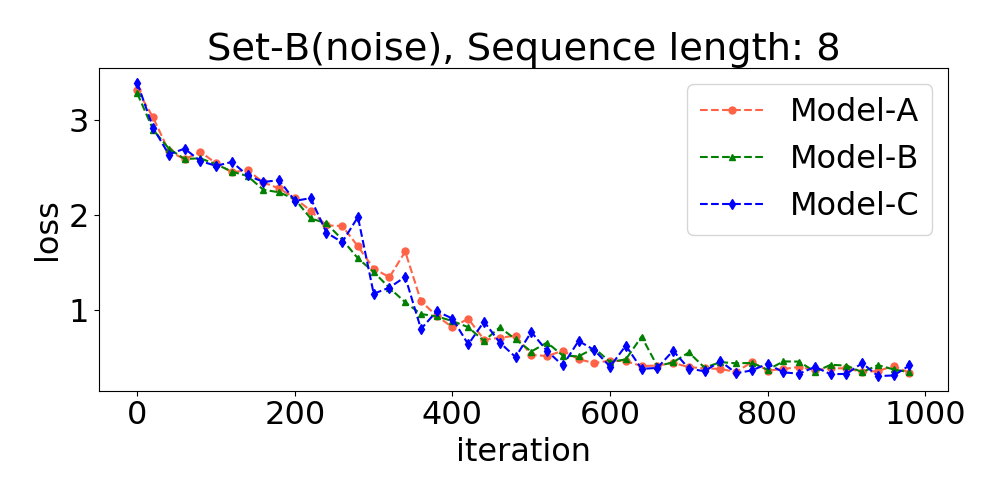} &
			\includegraphics[width=0.45\linewidth,keepaspectratio=true]{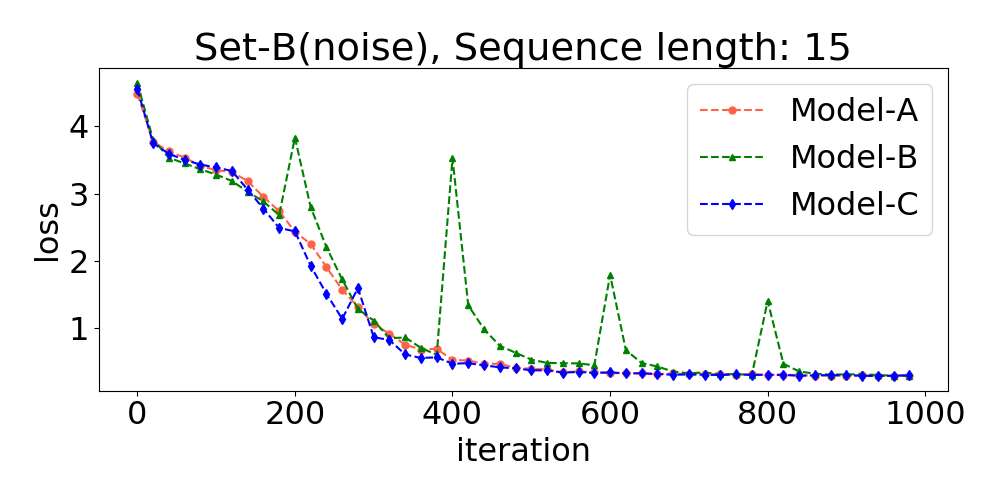} \\
            \includegraphics[width=0.45\linewidth,keepaspectratio=true]{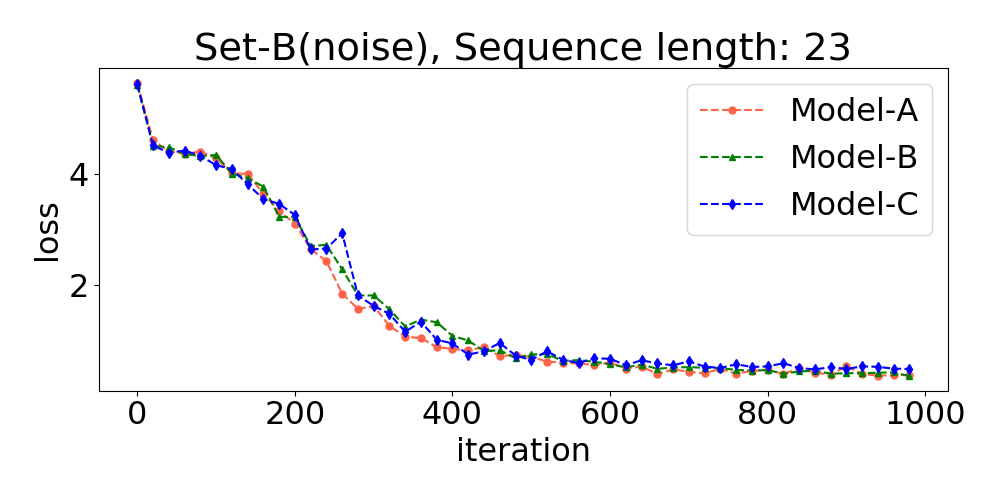} &
			\includegraphics[width=0.45\linewidth,keepaspectratio=true]{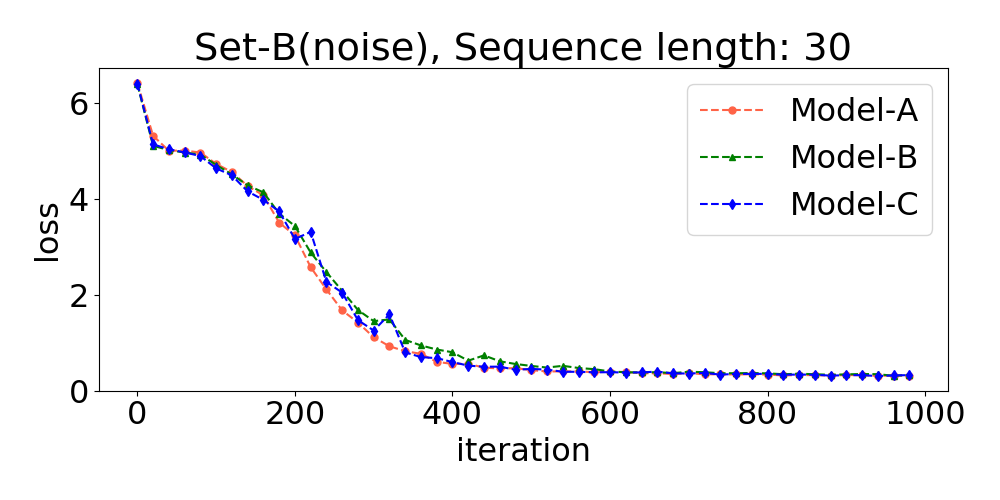} \\
            \includegraphics[width=0.45\linewidth,keepaspectratio=true]{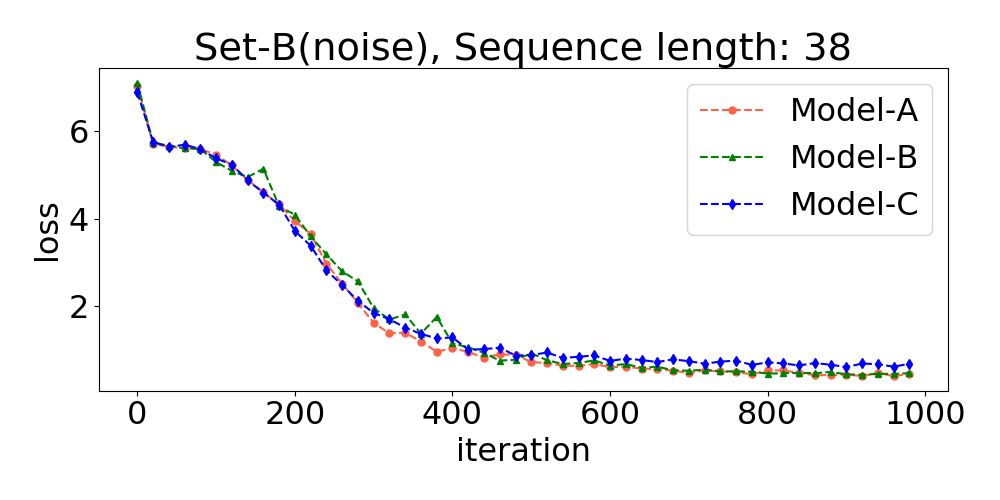} &
			\includegraphics[width=0.45\linewidth,keepaspectratio=true]{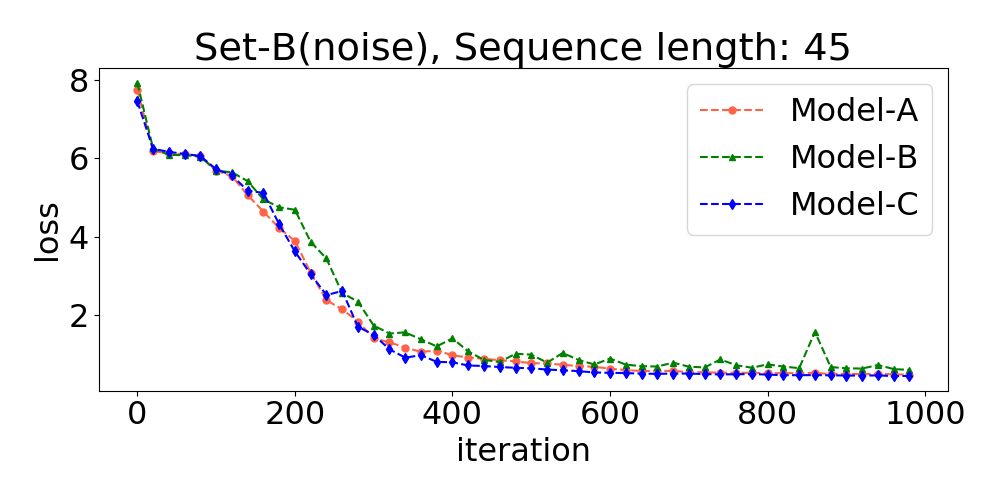} \\
		\end{tabular}
	\end{center}
    \caption{Loss graph of training process with subset 'Noise' of the Set-B.}
    \label{fig:mfcc_classes}
\end{figure}

\begin{figure}[h]
	\small
    \begin{center}
		\begin{tabular}{c c}
			\includegraphics[width=0.45\linewidth,keepaspectratio=true]{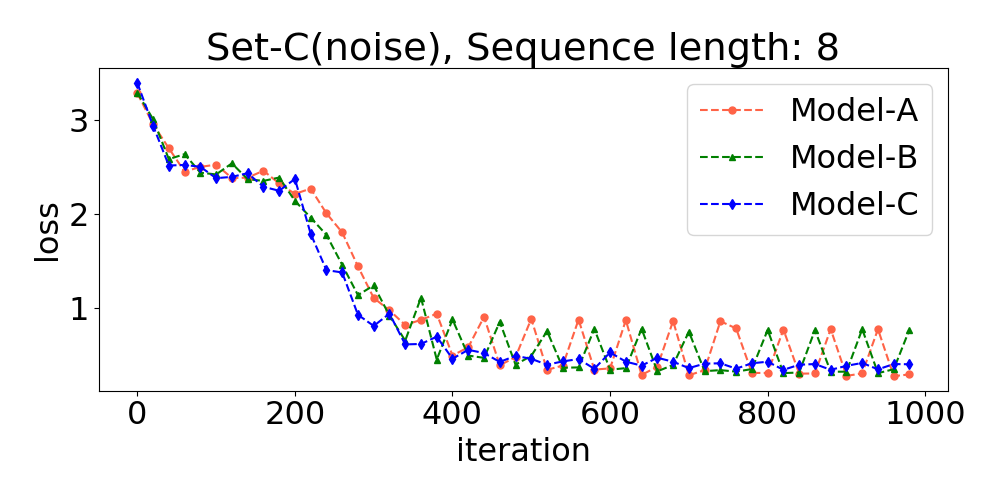} &
			\includegraphics[width=0.45\linewidth,keepaspectratio=true]{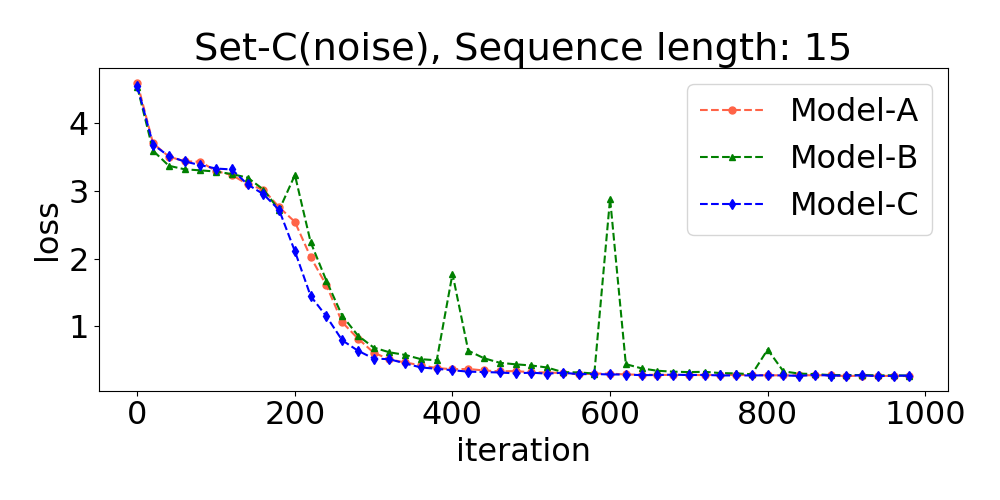} \\
            \includegraphics[width=0.45\linewidth,keepaspectratio=true]{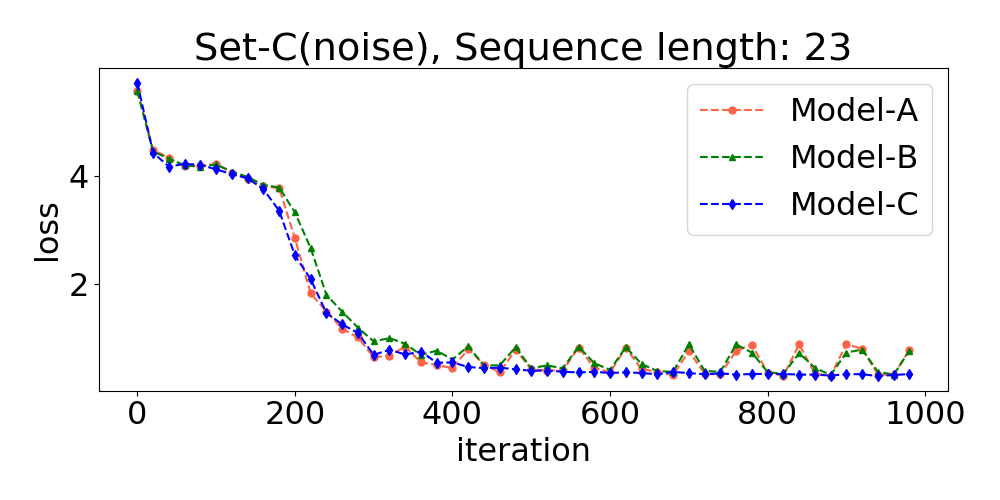} &
			\includegraphics[width=0.45\linewidth,keepaspectratio=true]{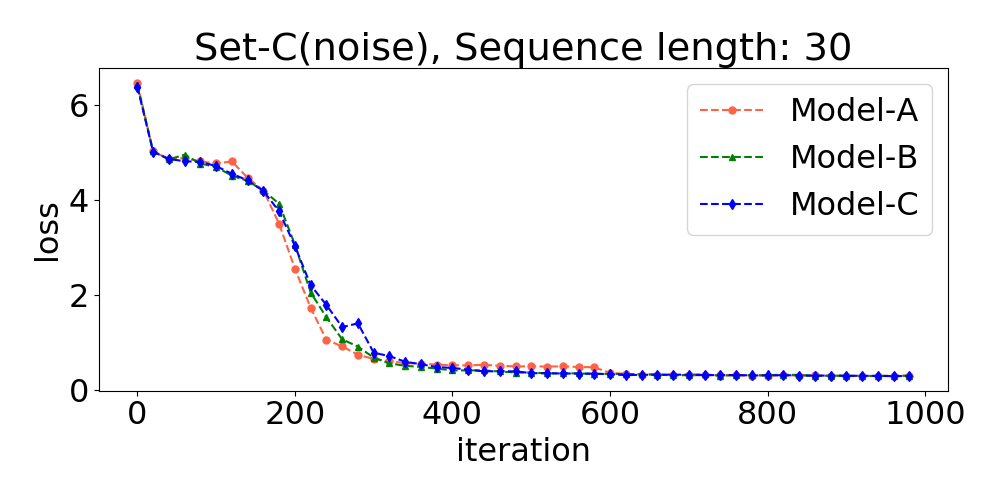} \\
            \includegraphics[width=0.45\linewidth,keepaspectratio=true]{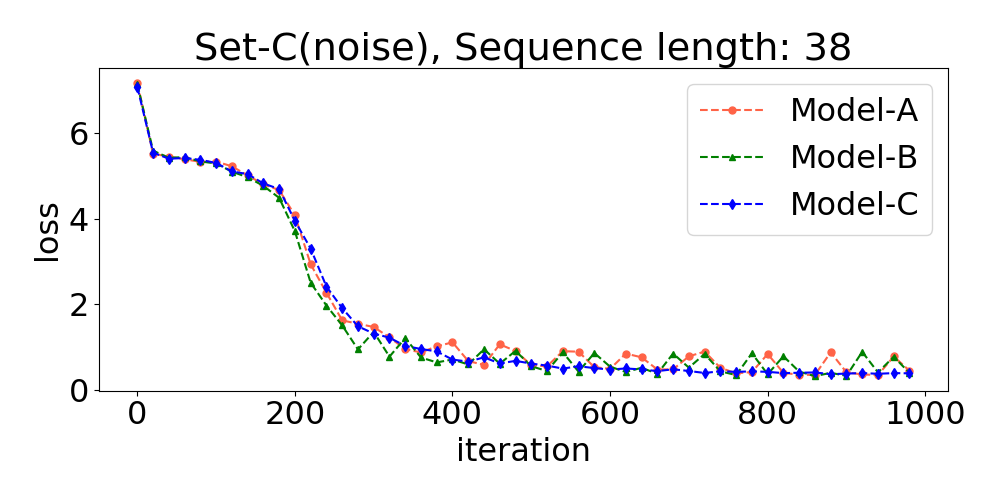} &
			\includegraphics[width=0.45\linewidth,keepaspectratio=true]{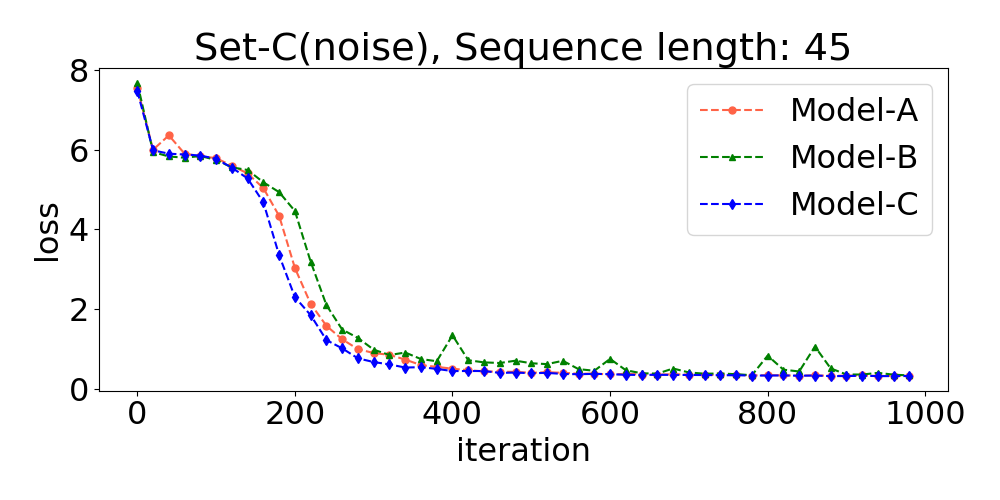} \\
		\end{tabular}
	\end{center}
    \caption{Loss graph of training process with subset 'Noise' of the Set-C.}
    \label{fig:mfcc_classes}
\end{figure}

\end{document}